%% file: egpaper_for_review.tex
\documentclass[10pt,twocolumn,letterpaper]{article}

\usepackage{iccv}
\usepackage{times}
\usepackage{epsfig}
\usepackage{graphicx}
\usepackage{amsmath}
\usepackage{amssymb}

\usepackage{soul}
\usepackage{units}
\usepackage{booktabs}
\usepackage{color}
\usepackage{bm}
\usepackage{mathtools}
\usepackage{diagbox}
\usepackage{tabulary}
\usepackage{makecell}
\usepackage{verbatim} 
\usepackage{dblfloatfix}
\usepackage{array,booktabs,ragged2e}
\usepackage{subfigure}
\usepackage[percent]{overpic}
\usepackage{realboxes}
\usepackage{makecell}
\usepackage{wrapfig}
\usepackage{slashbox}
\usepackage{enumitem}

\input{our-commands}

\usepackage[pagebackref=true,breaklinks=true,letterpaper=true,colorlinks,bookmarks=false]{hyperref}

\iccvfinalcopy 

\def\iccvPaperID{4341} 

\ificcvfinal\fi
\begin{document}

\title{Gaze360: Physically Unconstrained Gaze Estimation in the Wild}

\author{Petr Kellnhofer$^{*1}$,
Adri\`a Recasens$^{*1}$,
Simon Stent$^2$,
Wojciech Matusik$^1$,
and
Antonio Torralba$^1$\\
\vspace*{12pt}
$^1$ Massachusetts Institute of Technology, Cambridge MA 02139, USA \\
$^2$ Toyota Research Institute, Cambridge, MA, 02139, USA \\
{\tt\small \{pkellnho,recasens,wojciech,torralba\}@csail.mit.edu} \ \ {\tt\small simon.stent@tri.global} \\
{\small $*$ indicates equal contribution}
}

%
%

\maketitle

\begin{abstract}
   Understanding where people are looking is an informative social cue. In this work, we present \emph{Gaze360}, a large-scale gaze-tracking dataset and method for robust 3D gaze estimation in unconstrained images. 
   Our dataset consists of 238 subjects in indoor and outdoor environments with labelled 3D gaze across a wide range of head poses and distances.
   It is the largest publicly available dataset of its kind by both subject and variety, made possible by a simple and efficient collection method. 
   Our proposed 3D gaze model extends existing models to include temporal information and to directly output an estimate of gaze uncertainty.
   We demonstrate the benefits of our model via an ablation study, and show its generalization performance via a cross-dataset evaluation against other recent gaze benchmark datasets.
   We furthermore propose a simple self-supervised approach to improve cross-dataset domain adaptation.
   Finally, we demonstrate an application of our model for estimating customer attention in a supermarket setting. Our dataset and models are available at \url{http://gaze360.csail.mit.edu}.
\end{abstract}

\section{Introduction}
\label{sec:intro}

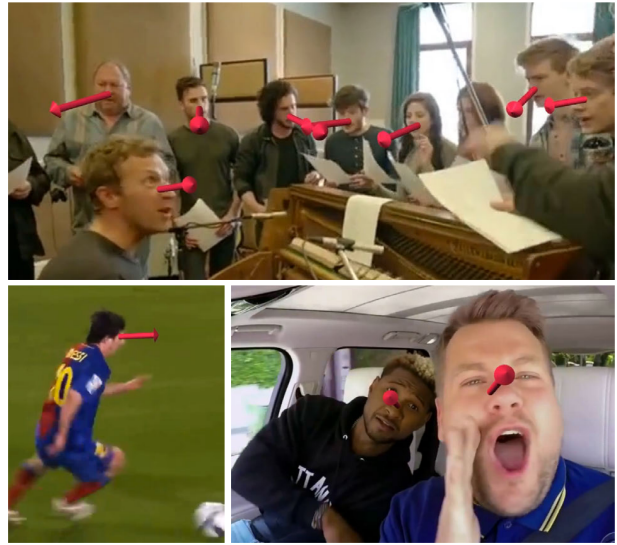
\myfigure{pull_figure}{\textbf{Overview}: we introduce a novel dataset and method for estimating 3D gaze in-the-wild. This figure illustrates our model's output on unseen video gathered from YouTube, demonstrating its robustness to diverse, physically unconstrained scenes.} 

In order to better understand humans -- their desires, intents and states of mind -- one must be able to observe and perceive certain behavioral cues.
Eye gaze direction is one such cue: it is a strong form of non-verbal communication, signalling engagement, interest and attention during social interactions~\cite{argyle1972non}.
Detecting and following where another person is looking is a skill developed early on in a child's life -- four-month-old infants are known to use eye gaze cuing to help visually process objects, for example~\cite{striano2006social}.
Just as a parent's gaze can help to guide a child's attention, human gaze fixations have also been found to be useful in helping machines to learn or interact in various contexts~\cite{palinko2016robot,sugano2016seeing}.

In recent years, while methods for related human modeling problems such as 2D body pose and face tracking have achieved impressive success by leveraging the representational power of deep convolutional neural networks along with very large annotated datasets~\cite{cao2017realtime,Guler2018DensePose,Hu_2017_CVPR,kocabas18prn,wei2016cpm}, methods for gaze estimation have not yet reached such levels of performance. 
This is primarily due to the lack of sufficiently large and diverse annotated training data for the task. Collecting precise and highly varied gaze data with ground truth, particularly outside of the lab, is a challenging task.


In this work, we introduce an approach to help tackle this task and narrow the perceived performance gap:
\begin{itemize}[noitemsep,topsep=0pt]
     \item we first describe a methodology to efficiently collect annotated 3D gaze data in arbitrary environments;
     \item we use our method to acquire the largest 3D gaze dataset in the literature by subject and variety, capturing video of 238 subjects in indoor and outdoor conditions, and we carefully evaluate the error and characteristics of the dataset;
     \item we train a variety of 3D gaze estimation models on the dataset before converging on a final model which uniquely takes a multi-frame input (to help resolve single frame ambiguities) and employs a pinball regression loss for error quantile regression to provide an estimate of gaze uncertainty;
     \item we demonstrate the usefulness of our dataset versus existing datasets by means of a cross-dataset model performance comparison (training on one dataset and testing on another), and introduce a simple method for self-supervised domain adaptation of gaze models;
     \item finally we demonstrate how our Gaze360 model can be applied to real-world use cases, such as estimating a customer's focus of attention in a supermarket.
\end{itemize}

\section{Related Work}
\label{sec:related}

\paragraph{Gaze datasets.}
A summary of comparable gaze datasets is shown in Table~\ref{tbl:Datasets}.
While many gaze-related datasets have been published in recent years~\cite{Huang2017,Khosla2016,mcmurrough2012eye,mora2014eyediap,smith2013gaze,sugano2014learning,weidenbacher2007comprehensive,zhang15_cvpr,Zhang2017}, they are mostly geared towards physically constrained applications such as desktop or smartphone gaze tracking.
Typically, these datasets are captured using a static recording setup \cite{mora2014eyediap,smith2013gaze,weidenbacher2007comprehensive,Zhu2017} or a camera integrated in a smartphone~\cite{Huang2017,Khosla2016,xu2015turkergaze}.
The static approach allows for more control and higher accuracy but can lack the diversity in illumination and motion blur useful for more general applications.
Smartphone-based solutions overcome these flaws and have the advantage of straightforward scaling via crowd-sourcing to increase the subject variety. However, they lack head pose and gaze variability due to the collocation of the device's camera and screen, as well as the screen's relatively narrow area for projecting targets.

To try to capture the nature of human gaze in arbitrary natural scenes, it is important not to overly constrain the subject's pose, allowing for coverage over the full gamut of head and eyeball orientations in relation to the camera.
While some existing datasets have relatively small head pose and gaze variation \cite{mcmurrough2012eye,mora2014eyediap,smith2013gaze}, others do provide a wider range \cite{Khosla2016,weidenbacher2007comprehensive,Zhu2017} but are still restricted to primarily frontal rather than oblique views. While it is true that the eyes become increasingly occluded at larger angles of head yaw, we wish to capture such cases so that our model can be used in less constrained settings.

In one of the most comprehensive datasets from Zhu~and~Deng~\cite{Zhu2017}, the authors increased acquisition speed and viewpont variety by using an array of cameras in different poses. However, the setup was restricted to collecting data in the lab environment. While our approach also uses a multi-camera setup, our goal was to quickly acquire many subjects at once, using a free-moving rather than fixed target that allowed us to capture the full range of gaze directions, as described in~\refFig{DataDistribution} and Section~\ref{sec:datasetsummary}. Moreover, as our capture setup is mobile, this allowed us to efficiently collect data from a broad demographic in more varied natural lighting environments, including a wider range of scale variation and image blur from subject motion during capture. This more closely approximates the domains of systems such as interactive robots or surveillance/monitoring cameras which might benefit from our gaze tracking model.

A recent work which also addresses gaze estimation in natural settings with larger camera-subject distances and less constrained subject motion, is that of~\cite{fischer2018rt}. Their approach to dataset generation was target-free, but required subjects to wear gaze-tracking glasses, used motion capture cameras to recover head pose, and needed a complicated semantic in-painting step to remove the gaze tracking glasses from the target image. In comparison, our approach is relatively simple, allowing us to scale to many more subjects (238 versus 15) and lighting conditions. 


\textbf{Geometric gaze models:}
Geometric models often use corneal reflections of near infra-red light sources \cite{hennessey2006single,yoo2005novel,zhu2005eye} or other light sources with known geometry \cite{Huang2017} to fit a model of the eyeball from which gaze can be inferred.
Since these methods rely on a physical model, they generalize quite easily to new subjects with little or no training data, but at the cost of higher sensitivity to input noise such as partial occlusions or lighting interference. Since they also rely on a fixed light source, they are not feasible in unconstrained settings such as ours.

\begin{table}[]
\centering
\caption{{\textbf{A comparison of popular gaze datasets.} The type and range of gaze labels, number of subjects and completeness of image data publicly available. \protect\emph{Full} stands for full face images, \protect\emph{Eyes} denotes crops of eye regions and \textit{N/A} means that the dataset was not available for use. Asterisks indicate datasets containing partially occluded face images.}}
\vspace{5pt}
\footnotesize
\label{tbl:Datasets}
\begin{tabular}{lcrrcc}
\hline
Dataset & Gaze & Range & \# Subj. & Image & Outdoor \\
\hline
TabletGaze~\cite{Huang2017TabletGaze} & \textcolor{red}{2D} & \textcolor{red}{$\sim$80\degrees} & \textcolor{orange}{51} & \textcolor{orange}{Eyes}  & \textcolor{red}{No}\\
iTracker~\cite{Khosla2016} & \textcolor{red}{2D} & \textcolor{orange}{$\sim$100\degrees} & \textcolor{green}{1,450} & \textcolor{green}{Full}  & \textcolor{orange}{Partially}\\
UT MV~\cite{sugano2014learning} & \textcolor{green}{3D} & \textcolor{red}{$\sim$50\degrees} & \textcolor{orange}{50} & \textcolor{orange}{Eyes}  & \textcolor{red}{No}\\
Columbia~\cite{smith2013gaze} & \textcolor{green}{3D} & \textcolor{red}{60\degrees} & \textcolor{orange}{56} & \textcolor{orange}{Full*} & \textcolor{red}{No}\\
RT-GENE~\cite{fischer2018rt} & \textcolor{green}{3D} & \textcolor{red}{75\degrees} & \textcolor{red}{15} & \textcolor{orange}{Full*}  & \textcolor{red}{No}\\
MPIIFaceGaze~\cite{Zhang2017} & \textcolor{green}{3D} & \textcolor{red}{$\sim$80\degrees} & \textcolor{red}{15} & \textcolor{green}{Full}  & \textcolor{red}{No}\\
EYEDIAP~\cite{mora2014eyediap} & \textcolor{green}{3D} & \textcolor{orange}{90\degrees} & \textcolor{red}{16} & \textcolor{green}{Full} & \textcolor{red}{No}\\
Weidenb.~\cite{weidenbacher2007comprehensive} & \textcolor{green}{3D} & \textcolor{orange}{180\degrees} & \textcolor{red}{20} & \textcolor{red}{N/A} & \textcolor{red}{No} \\
Zhu~\cite{Zhu2017} & \textcolor{green}{3D} & \textcolor{orange}{180\degrees} & \textcolor{green}{200} & \textcolor{red}{N/A}  & \textcolor{red}{No}\\
\textbf{Gaze360~[ours]} & \textbf{\textcolor{green}{3D}} & \textbf{\textcolor{green}{360\degrees}} & \textbf{\textcolor{green}{238}} & \textbf{\textcolor{green}{Full}}  & \textbf{\textcolor{green}{Yes}} \\
\hline
\end{tabular}
\end{table}

\textbf{Appearance-based gaze models:} Appearance-based methods learn a more direct image-to-gaze mapping, using large datasets of annotated eye or face images.
Support vector regression \cite{xu2015turkergaze}, random forests \cite{Huang2017TabletGaze} and most recently deep learning \cite{fischer2018rt,Khosla2016, zhang15_cvpr,Zhang2017, Zhu2017} have been applied in this way.
A preprocessing step of eye or face detection is often required \cite{Khosla2016,zhang15_cvpr}. Our model does not rely on eye or face detectors, which enables it to achieve higher robustness in unconstrained settings when the required features become partially occluded.
Dependency between gaze and head pose can either be handled by training implicitly \cite{Khosla2016,zhang15_cvpr,Zhang2017} or modeled explicitly with separate branches \cite{Zhu2017}.

Gaze estimation becomes more difficult under partial occlusion of eyes. Even at $90-135$\degrees~head yaw a significant part of one eyeball is often still visible and informative for gaze estimation (see Supplemental). 
Existing methods~\cite{Khosla2016,zhang2019mpiigaze} do not deal with these cases and typically assume that the subject is facing the camera. However, such models do not generalize well to challenging applications such as in robotics or surveillance. 
Unlike previous approaches, our model is designed to cope with such situations by always providing best effort prediction along with an appropriate confidence measure. 
We learn to predict uncertainty via quantile regression~\cite{koenker_2005} learned using a pinball loss.
Our model outputs an estimated gaze direction even with fully occluded eyes by relying on visible head features, while at the same time informing about the limited accuracy of its prediction by outputting a correspondingly higher uncertainty value.
In addition, unlike previous models, we investigate the use of additional frames to improve gaze estimates through the aggregation of image evidence over time. This increases the chance of capturing relevant features that may only be visible in few frames. We show how using motion significantly helps the system performance over a wide range of view angles. 



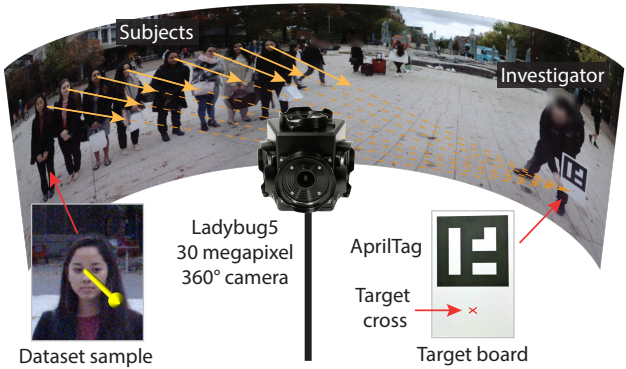
\myfigure{panorama}{\textbf{Acquisition setup}. Our setup allows us to efficiently collect large volumes of diverse, annotated data for 3D gaze estimation. We create a dataset with 238 subjects in a wide range of lighting conditions (both indoor and outdoor) and distances and angles to subjects.} 

\section{Dataset collection method}
\label{sec:dataset}

There is currently no dataset suitable to learn a model capable of robustly estimating 3D gaze in-the-wild. 
Previous efforts to record large-scale datasets relied on careful acquisition setups with precisely measured subject and gaze target positioning~\cite{mora2014eyediap,sugano2014learning,Zhu2017}. Such setups are nearly impossible to move to different locations, can only record single subjects at a time and require constant verification of the desired gaze from the subject which makes the collection process inflexible and very slow. This is the reason why all existing datasets with 3D gaze labels are recorded in indoor environments and frequently use few subjects. 
As evidenced by the success of 2D body and face tracking models in the wild~\cite{cao2017realtime}, to improve in-the-wild robustness it is important to collect data with a large number of different subjects, large variation in natural illumination and a wide range of head poses and gaze directions. 

\subsection{Setup}
\label{sec:setup}

To tackle these issues we opted for a setup built around a Ladybug5 360\degrees~panoramic camera (\refFig{panorama}) placed on a tripod in the center of the scene, and a large moving rigid target board marked with an AprilTag~\cite{wang2016iros} and a cross on which subjects were instructed to continuously fixate.
This allowed data from multiple subjects to be recorded simultaneously. 
The Ladybug5 consists of five synchronized and overlapping 5 megapixel camera units each with 120\degrees~ horizontal field of view, plus one additional upward-facing camera which we do not use. We store each frame as $3382\times4096$ pixels image after fish-eye lens rectification.
The face of a subject standing one meter away from the camera could be fully captured in at least one of the views. The camera is factory-calibrated and we rectified all images after capture to remove barrel distortion. 
The compactness of the setup, consisting of a single camera unit on a tripod together with a laptop and portable power source, allowed for easy portability and deployment for efficient data collection in many environments.

\textbf{Subject positioning.}
To build the dataset, we use AlphaPose~\cite{fang2017rmpe} to detect the position of head keypoints and feet of subjects in rectified frames from each camera unit independently. For very close subjects whose feet are beyond the camera field of view, we use the average body proportions of standing subjects to estimate their feet position from their hip position.
The Ladybug camera provides a 3D ray in a global Ladybug Cartesian coordinate system $L = [\mathbf{L}_x,\mathbf{L}_y,\mathbf{L}_z]$ for every image pixel. We use it to derive the position of feet and eyes in spherical coordinates. The remaining unknown variable is the distance from Ladybug origin to eyes, $d$. We exploit a measured camera height above the horizontal ground plane that the camera and all subjects stand on. Although this limits our training data collection to flat surfaces, it is not restrictive at test-time. For further details on the trigonometry, please consult the supplementary materials.

\textbf{Target positioning:} Our target consists of a white board with a large AprilTag~\cite{wang2016iros} on one side and a smaller cross beside it on both sides (\refFig{panorama}). The cross serves as a gaze fixation target for the study subjects while the tag is used for tracking of the board in 3D space.
We use the original AprilTag library to detect the marker in each of the camera views and estimate its 3D pose using the known camera calibration parameters and marker size. We then use the pose and known board geometry to find the 3D location of the target cross $\mathbf{p}_t$.

\textbf{Gaze direction:} We compute the gaze vector in the Ladybug coordinate system as a simple difference $\mathbf{g}_L = \mathbf{p}_t - \mathbf{p}_e$. However, such a form would change with rotation of the camera and its coordinate system $L$. To remedy this, we express the gaze in the observing camera's Cartesian eye coordinate system $E = [\mathbf{E}_x,\mathbf{E}_y,\mathbf{E}_z]$. $E$ is defined so that the origin is $\mathbf{p}_e$, $\mathbf{E}_z$ has the same direction as $\mathbf{g}_L$ and $\mathbf{E}_x$ lies in a plane defined by $\mathbf{L}_x$ and $\mathbf{L}_y$ (no roll). We can then convert the gaze vector to the eye coordinate system by:
\begin{eqnarray}
\mathbf{g} = E \cdot \frac{\mathbf{g}_L}{||\mathbf{g}_L||_2}.
\end{eqnarray}
This definition of gaze direction guarantees that $\mathbf{g} = [0,0,-1]$ when the subject looks directly at the camera, independently of the subject's position, and in general allows to express the gaze orientation from the local appearance of the head without the need for any global context.

\subsection{Acquisition procedure}

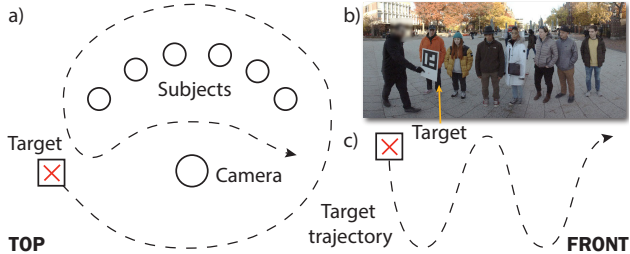
\myfigure{path}{\textbf{Dataset collection protocol}: (a) the top view of the scene and target board trajectory showing full coverage around the subjects; (b) the image of the scene from the camera (stitched for illustration only); (c) the side view of the scene and target board trajectory showing large induced variation in pitch to the target. \vspace{-0.5cm}}

We acquired an institution review board approval for our dataset collection experiment.
Subjects were instructed to stand around a camera at a distance of between $1-3$m (average $2.2$m) and continuously track the target cross on the side of the marker board visible to them (\refFig{path}).
For safety, subjects were instructed to stay approximately in their starting locations as they would not be able to both track the target and see possible obstacles while moving.

The marker board was manipulated by one of the investigators who carried it once in a large loop around both the subjects and the camera ($2-5$m radius) and then in between the camera and subjects (\refFig{path}a). While in motion, the target board was simultaneously moved up and down (\refFig{path}c) to elicit gaze pitch variation. The loop part of the trajectory allowed to cover all possible gaze directions. The inner path was added to sample more extreme gaze pitch variation which can only be achieved from a closer distance due to limitations on the vertical position of the marker in the scene. We ensured that the marker board was always positioned to face the camera with the AprilTag as fronto-parallel as possible to reduce pose estimation error (\refFig{path}b).

In order to capture a wide range of relative eyeball and head poses, we alternated between ``move'' and ``freeze'' instructions during each capture. While in the ``move'' state, subjects were allowed to naturally orient their head and body pose to help track the target. When the ``freeze'' instruction was issued, subjects were only allowed to move their eyes while maintaining a fixed head pose if possible.

\section{Gaze360 dataset summary}
\label{sec:datasetsummary}

Our dataset is unique for its combination of 3D gaze annotations, wide range of gaze and head poses, variety of indoor and outdoor capture environments and diversity of subjects. It is only surpassed in number of subjects by the GazeCapture~\cite{Khosla2016} dataset (1,450 subjects), which is 2D and covers only a narrow gaze range for a limited use case. See Table~\ref{tbl:Datasets} for a dataset comparison. Notably, our dataset is also the first to provide these qualities for short continuous videos (8\,Hz).


\paragraph{Summary statistics.}

We collected 238 subjects in 5 indoor (53 subjects) and 2 outdoor (185 subjects) locations over 9 recording sessions. 
This is an acquisition speed that is unmatched by other on-site techniques and can only be compared to crowd-sourced approaches which, however, cannot compete in terms of experimental control.
In total we acquired 129K training, 17K validation and 26K test images with gaze annotation. 
For privacy reasons we did not survey additional data about our subjects, but a visual inspection shows a wide distribution of subject ages, ethnicities and genders (58\,\% female, 42\,\% male). Please refer to \refFig{Dataset} for examples. 

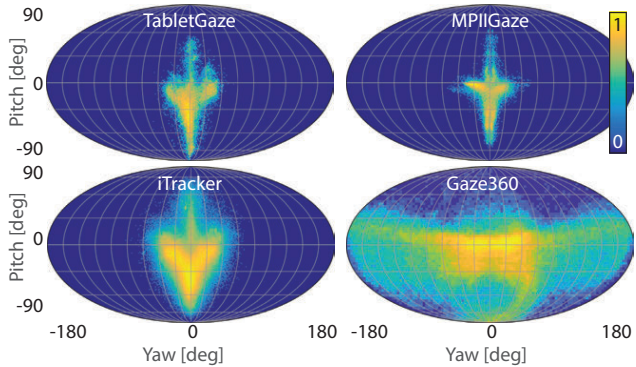
\myfigure{DataDistribution}{\label{fig:datadistribution}\textbf{Dataset statistics.} Joint distributions of the gaze yaw and pitch for TabletGaze~\cite{Huang2017}, MPIIFaceGaze~\cite{Zhang2017}, iTracker~\cite{Khosla2016} and our Gaze360 dataset. The Mollweide projection used to visualize the full unit sphere surface. All intensities are logarithmic.}

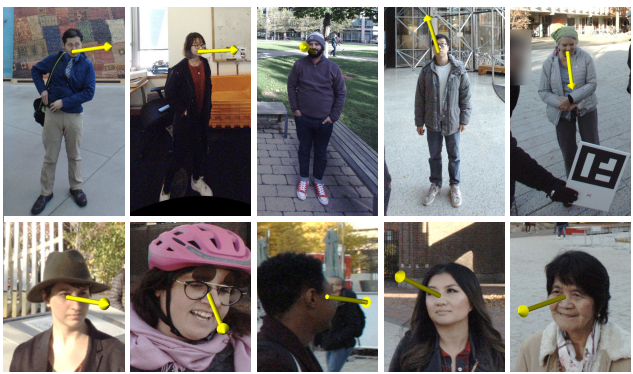
\myfigure{Dataset}{\textbf{Gaze360 dataset samples}: showing the diversity in environment, illumination, age, sex, ethnicity, head pose and gaze direction. Top: full body crops; bottom: closer-up head crops. Yellow arrows show measured ground-truth gaze.}


\paragraph{Data distribution.}
We plot the angular distribution of the gaze labels covered by our and several other datasets using the Mollweide projection in~\refFig{DataDistribution}. This illustrates how our dataset covers the entire horizontal range of 360\degrees. While a portion of these gaze orientations correspond to fully occluded eyes (facing away from the camera), our dataset allows for gaze estimation up to the limit of eye visibility. This limit can, in certain cases, correspond to gaze yaws of approximately $\mypm140$\degrees~(where the head pose is at 90\degrees~such that one eye remains visible, and that eye is a further 50\degrees~rotated).
The vertical range is limited by the achievable elevation of the marker. Sampling is less dense in the rear region (around the left and right borders of the map). This can be explained by occlusion of the target board by the subjects. 

\paragraph{Error characterization.}
%
In order to validate the accuracy of our gaze annotations we conducted a control experiment. We followed the standard acquisition procedure with our 360\degrees~camera and a single participant at a time wearing an additional front-facing test camera mounted above the right eye. We measured the 3D gaze in the test camera using the standard AprilTag based procedure and the known origin coinciding with the camera. Additional AprilTags in the background were used to register both cameras. We measured the mean difference between both gaze labels to be $2.9$\degrees~over three recordings of two subjects. This is well within the error of appearance-based eye tracking at distance, validating our acquisition procedure as a means of collecting an annotated 3D gaze dataset.

\section{Gaze360 model}
\label{sec:model}

Gaze is a naturally continuous signal. Gaze fixations and transitions yield a sequence of gaze directions. To exploit this, we propose a video-based gaze-tracking model using bidirectional Long Short-Term Memory capsules (LSTM)~\cite{graves2005bidirectional}, which provide a means of modeling sequences where the output for one element is dependent on both past and future inputs. In this paper, we utilize sequences of 7 frames to predict the gaze of the central frame. Note that other sequence lengths including a single central frame alone are also possible.

\refFig{model_figure} illustrates the architecture of the Gaze360 model. A head crop from each frame is individually processed by a convolutional neural network (backbone), which produces high-level features with dimensionality $256$. These features are fed to bidirectional LSTMs with two layers which digest the sequence within forward and backward vectors. Finally, these vectors are concatenated and passed through a fully connected layer to produce two outputs: the gaze prediction and an error quantile estimation.

The gaze prediction output regresses the angle of the gaze relative to the camera view. 
In previous work, 3D gaze was predicted as a unit gaze vector \cite{mora2014eyediap,Zhu2017} or as its spherical coordinates \cite{sugano2014learning,Zhang2017}. We use spherical coordinates which we believe to be more naturally interpretable in this context. 
We define the spherical coordinates such that the pole singularities correspond to strictly vertical gaze oriented either up or down, which are very rare directions. 

We use an ImageNet-pretrained ResNet-18~\cite{he2016deep} as the backbone network. All the models were trained in PyTorch using the Adam optimizer~\cite{kingma2014adam} with learning rate $10^{-4}$.

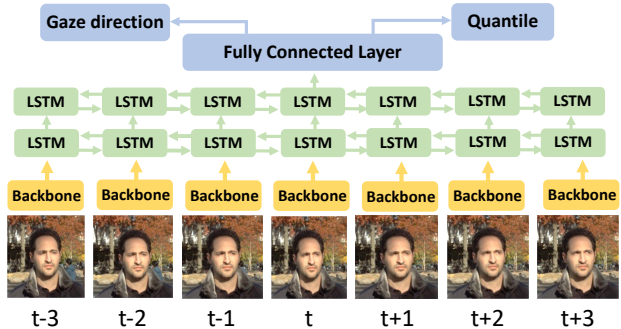
\myfigure{model_figure}{\textbf{Gaze360 model architecture}. The model receives multiple frames of input which are passed through a backbone network. The output for each frame is fed to a bidirectional LSTM to produce the compact representation which is used to make the final prediction of gaze direction and quantile regression. We use a 7-frame input window centered around the target frame.}

\subsection{Error quantile estimation}

To the best of our knowledge, all existing research applying neural networks to the task of gaze estimation do not consider error bounds. Error bounds are useful when estimating gaze in unconstrained environments, because precision is likely to degrade when the eye is viewed from a sideways angle, or when one or more eyes are partially obscured (e.g. by glasses frames). In a classification setting, softmax outputs are often used as a proxy for confidence. However, for regression this is not possible, as the magnitude of the output corresponds directly to the predicted property. 

To model error bounds, we use a pinball loss function~\cite{koenker_2005} to predict error quantiles. We use one single network to predict both the mean value and the $10\%$ and $90\%$ quantile. The effect of this is that for a given image, we estimate through a single forward pass both the expected gaze direction and a cone of error within which the ground truth should lie $80\%$ of the time.
We assume that the distribution is isotropic in our spherical coordinate system. This assumption is not strictly true, especially for large pitch angles due to the space distortion around pole singularities. However, for most of the observed gaze directions (\refFig{DataDistribution}) it is a reasonable approximation to reduce dimensionality and simplify the interpretation of the result.

The output of our network is $f(I)=(\theta,\phi,\sigma)$, where $(\theta,\phi)$ is the expected gaze direction in spherical coordinates, for which we already have a corresponding ground truth gaze vector in the eye coordinate system $\mathbf{g}$ (see Sec.~\ref{sec:setup}) as $\theta = -\arctan{\frac{g_x}{g_z}}$ and $\phi = \arcsin{g_y}$.
The third parameter, $\sigma$, corresponds to the offset from the expected gaze such that $\theta+\sigma$ and $\phi+\sigma$ are the $90\%$ quantiles of their distributions while $\theta-\sigma$ and $\phi-\sigma$ are $10\%$ quantiles.

Finally, we compute the pinball loss of this output. This will naturally force $\phi$ and $\theta$ to converge to their ground truth values and $\sigma$ to the quantile threshold. If $y = (\theta_{\textrm{gt}},\phi_{\textrm{gt}})$, the loss $L_\tau$ for the quantile $\tau$ and the angle $\theta$ can be written as:
\begin{eqnarray}
    \hat{q}_\tau &=&  
\begin{cases}
    \theta_{\textrm{gt}}-(\theta-\sigma),& \text{for } \tau \leq 0.5\\
    \theta_{\textrm{gt}}-(\theta+\sigma),              & \text{otherwise}
\end{cases} \\
L_\tau(\theta,\sigma,\theta_{\textrm{gt}}) &=& \textrm{max}(\tau\hat{q}_\tau,-(1-\tau)\hat{q}_\tau).
\end{eqnarray}
A similar formulation is used for the angle $\phi$.
We average the losses for both angles and quantiles $\tau=0.1$ and $\tau=0.9$. Thus, $\sigma$ is a measure of the difference between the $10\%$ and $90\%$ quantiles and the expected value.

\subsection{Adapting to unseen domains}
\label{subsec:unseen}
Despite the variety in the Gaze360 dataset, some real-world applications may benefit from a closer adaptation of the model to the target domain. For this reason, we introduce a self-supervised method for domain adaptation. 

Our general model is fine-tuned using a mix of the labeled Gaze360 images and unlabeled images from the new domain.
Inspired by \cite{tzeng2017adversarial}, we introduce a discriminator which tries to identify the source domain of the image features as a binary classification task. The features are the output of the backbone network. The discriminator loss $L_D$ is added to the original supervised loss $L_\tau$ for those images where ground truth is available.

In addition, we added a further loss to exploit the left-right symmetry of the gaze-estimation task as a means of encouraging model output consistency on unlabeled data. We use the model to compute the gaze of the original and horizontally flipped image, and the pinball loss $L_S$ to minimize the angular difference between the prediction from the first input and horizontally mirrored prediction from the second input. While this loss by itself can lead to collapse to a gaze prediction along the line of symmetry, our observations in \refSec{res_cross} show that this helps when used as a regularizer to improve performance in an unseen target domain. Altogether we minimize $L = \alpha \cdot L_\tau + L_D + \beta \cdot L_S$ where $\alpha = 60$ and $\beta = 3$ in our experiments.

\section{Experimental Analysis}
\label{sec:experiments}

\subsection{Model evaluation}

In this section, we compare several approaches using the Gaze360 dataset. 
We compared the following methods: 
\textbf{Mean} - uses the mean gaze of the training set for all predictions;
\textbf{Deep Head Pose} - a deep network based head pose estimator by Ruiz~\etal \cite{ruiz2017deephp};
\textbf{Static} - the backbone model, ResNet-18, and two final layers to compute the prediction;
\textbf{TRN} - a version of Temporal Relation Network~\cite{zhou2018temporal} where the features of frames at fixed windows around time $t$ are concatenated before averaging the predictions of the temporal windows;
\textbf{LSTM} - refers to the \textbf{Gaze360} architecture. 

For each of the three architectures introduced above, we report accuracy of different baselines for uncertainty estimation:
\textbf{MSE} -  uses the mean squared error to regress only the spherical angles of gaze without uncertainty;
\textbf{MSE+Drop} - using the MSE model, the uncertainty is estimated by 5 forward passes for each input while randomly dropping neurons in the last layer and computing the variance of the output;
\textbf{Crop augmentation} - $5$ random head crops are sequentially evaluated to estimate uncertainty using the variance of the $5$ predictions of the MSE-trained model; and
\textbf{Pinball Loss} - gaze direction and error bounds are jointly estimated using the pinball loss. 

The angular errors in Table~\ref{tab:results} are provided separately for the entire test set (\emph{All 360\degrees}) and for samples where the subject is looking within $90$\degrees~(\emph{Front 180\degrees}) and $20$\degrees~(\emph{Front facing}) of the camera direction. We also report the Spearman's rank correlation between the error quantile estimate and the actual error, which is a metric for how well the predicted error bounds estimate the actual error. 

The results confirm that eye-free \textbf{Mean} predictions as well as \textbf{Head pose} are insufficient to predict the rich variation of eye movement in our dataset. All of our gaze models outperform these simple baselines. We also observe that, under the same conditions, the error is generally lowest for the model using \textbf{Pinball} loss. The same trend can be seen for the correlation between the predicted uncertainty and actual prediction error. Additionally, only a single forward pass is required for the prediction. Hence, we chose the \textbf{Pinball} loss as our recommended approach.
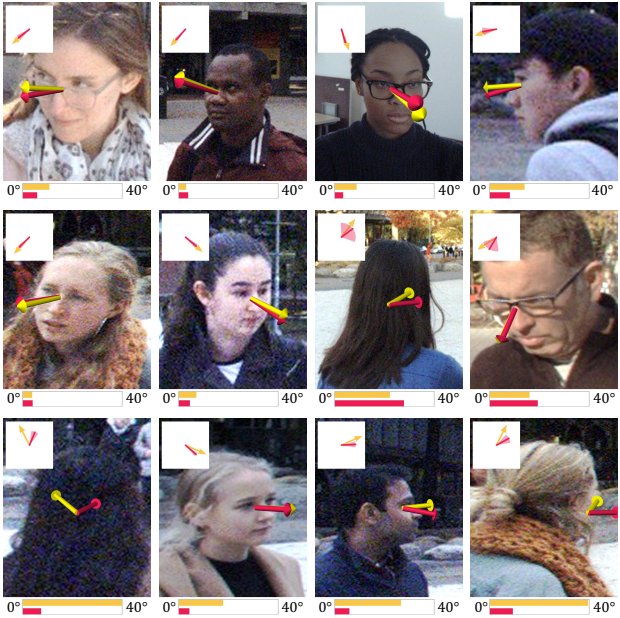
\myfigure{Results}{\textbf{Test set examples}: ground truth gaze (yellow) and Gaze360 predictions (red) are shown for unseen test subjects. The bars denote actual (yellow) and predicted (red) errors in degrees. The inset shows a top-down view of the gaze estimates and the predicted error versus ground truth. The bottom row shows sample failure cases where the model was overconfident. }

Switching from a single-frame static model to a temporal model also benefits the gaze prediction accuracy substantially. We conclude that although the performance of \textbf{TRN} and \textbf{LSTM} is similar, we recommend the \textbf{Pinball LSTM} for its slightly better results in our metric and straightforward adaptation to use a different number of input frames. 

\begin{table}[t]
\centering
\caption{\textbf{Performance comparison on Gaze360 dataset.} The table below reports the mean angular errors for various models and benchmarks on the Gaze360 test data. The last column shows the correlation between the actual error and the predicted uncertainty.}
\vspace{5pt}
\footnotesize
\begin{tabular}{lccccc}
\hline
\textbf{Model} & \thead{Uncert. \\ Loss} & \emph{\thead{All \\ 360\degrees}}     &\emph{\thead{Front \\ 180\degrees}}  & 
\emph{\thead{Front \\ Facing}} & \thead{Uncert. \\ Corr.}  \\ \hline
\textbf{Mean}  & - & 59.0 & 40.5 & 19.0  & - \\ \hline
\textbf{Deep HP} & - & 49.3 & 30.7 & 22.7 & - \\ \hline

\textbf{MSE Static} & No & 15.8 & 13.7 & 13.4 & - \\ 
\textbf{MSE TRN} & No & 14.3 & 11.8 & 11.8 & - \\ 
\textbf{MSE LSTM} & No & 14.1 & 12.1 & 11.6 & - \\ \hline

\textbf{MSE+Drop Static}  & No & 15.8 & 13.7 & 13.4 & 0.24 \\
\textbf{MSE+Drop TRN}  & No & 14.3 & 11.8 & 11.8 & 0.31 \\
\textbf{MSE+Drop LSTM}  & No & 14.1 & 12.1 & 11.6 & 0.31 \\ \hline

\textbf{Crop Aug. Static} & No &  16.0 & 13.2 & 12.6 & 0.37 \\  
\textbf{Crop Aug. TRN} & No & 14.2 & 11.5 & 11.4 & 0.39 \\ 
\textbf{Crop Aug. LSTM} & No & 14.1 & 11.6 & 11.2 & 0.37 \\ \hline

\textbf{PinBall Static} &  Yes & 15.6 & 13.4 & 13.2 & 0.42 \\ 
\textbf{PinBall TRN} &  Yes & 14.1 & 11.7 & 11.6 & \textbf{0.46} \\
\textbf{\makecell[l]{Pinball LSTM \\(i.e., Gaze360)}} &  Yes & \textbf{13.5} & \textbf{11.4} & \textbf{11.1} & 0.45 \\ \hline

\end{tabular}
\label{tab:results}
\end{table}

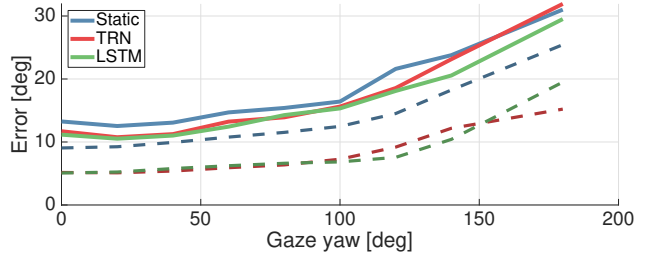
\myfigure{PlotYaw}{\textbf{Error measured on Gaze360 dataset using the \textbf{Pinball} models.} The full lines show prediction error, the dashed lines show predicted uncertainty.\vspace{-0.4cm}}

In \refFig{PlotYaw} we present the prediction error of the models using \textbf{Pinball} loss as a function of gaze yaw angle. As expected, accuracy falls with increasing gaze yaw angle. Unlike traditional eye trackers, our model smoothly transitions into head pose estimation (between head yaws of 90-150\degrees) to provide a best guess of gaze even for rear views. This is accompanied by a higher associated uncertainty (dashed lines). Although the error for frontal views is generally larger than errors reported on existing high-resolution datasets, we next show that this is due to the challenging properties of Gaze360 which allow models trained on it to transfer better to physically unconstrained images.

In \refFig{Results} we show sample results on our test data. The angular error denoted by the yellow bar intuitively grows as the eyes become smaller due to distance or occluded due to head pose variation. Although the prediction error for away-looking poses is on average large, the uncertainty measure provides a reasonable prediction of this behavior.

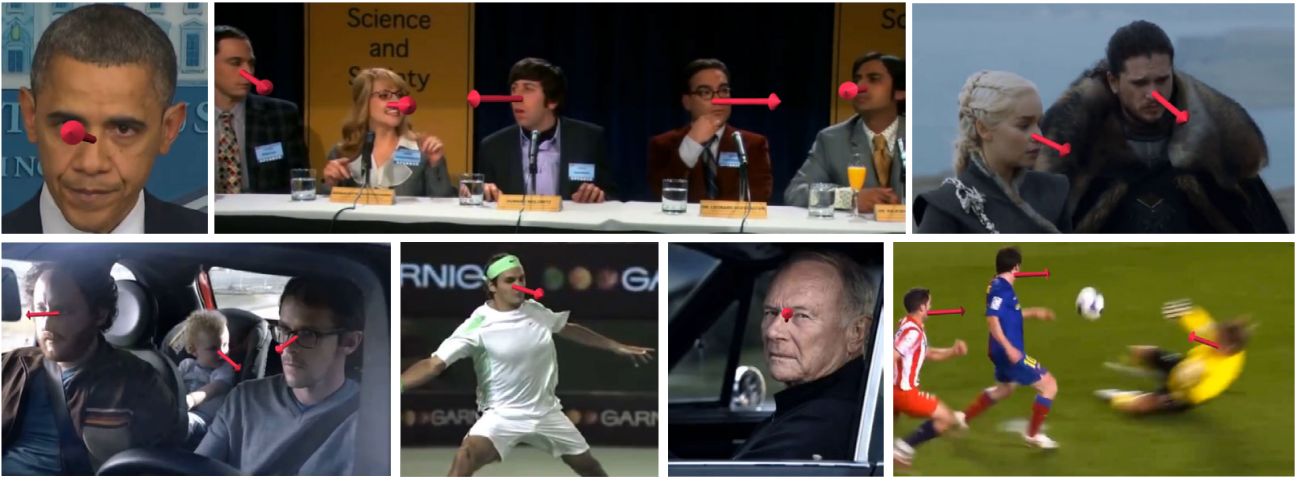
\mycfigure{youtube_figure}{\textbf{Estimating 3D gaze in the wild}: further examples of our model's output on unseen video gathered from YouTube.}

\begin{table}[b!]
\centering
\caption{\textbf{Cross-dataset evaluation:} we report the mean angular errors for the Static model trained using different datasets.}
\vspace{5pt}
\footnotesize
\begin{tabular}{l|c|c|c|c}
\backslashbox{\textbf{Train}}{\textbf{Test}} & \textbf{Columbia}  & \textbf{\thead{MPII\\FaceGaze}}    &\textbf{\thead{RT-GENE}} & \textbf{Gaze360} \\ \hline
\textbf{Columbia}     & -    & 12.3 &32.8 &57.9 \\ 
\textbf{MPIIFaceGaze} & 12.4 & -     &26.5& 57.8 \\ 
\textbf{RT-GENE} & 24.2 & 18.9 & - & 56.6 \\ 
\textbf{Gaze360}     &  9.0 & 12.1  & 23.4 &  - \\ 
\textbf{Gaze360 + DA}     &  \textbf{8.1} & \textbf{9.9} & \textbf{21.9} &  - \\ 
\end{tabular}
\label{tab:cross}
\end{table}

\subsection{Cross-dataset evaluation}
\label{sec:res_cross}
We evaluate the value of the Gaze360 dataset for gaze estimation in the wild by training the \textbf{Pinball Static} model using multiple pre-existing 3D gaze datasets and measuring cross-dataset test error.
The comparison datasets we use are:
\textbf{Columbia}~\cite{smith2013gaze} - high-resolution close-up faces;
\textbf{MPIIFaceGaze}~\cite{Zhang2017} - faces captured by webcams;
\textbf{RT-GENE}~\cite{fischer2018rt} - low-resolution faces using in-painting to mask out eye-tracking glasses;
\textbf{Gaze360} (Ours) - faces with varying resolution;
For those datasets where no official splits were provided~\cite{smith2013gaze,Zhang2017} we use all available samples for training and do not measure the within-domain error.

Table~\ref{tab:cross} summarizes the results. This task is much more challenging than within-domain tests. The best results are consistently achieved when our dataset is used for training.
In addition, we fine-tune our Gaze360-trained model on new domains (\textbf{Gaze360 + DA}) using the self-supervised approach described in Sec.~\ref{subsec:unseen}, which does not utilize the ground truth labels in other datasets. 
Our domain adaption strategy improves performance further on all the datasets. 

\section{Tracking gaze in the wild}
\textbf{Prediction in unconstrained environments}: The variation in appearance of subjects in the Gaze360 dataset allows our model to perform well without further training or fine-tuning on unseen image and video data from uncurated online sources. We demonstrate this visually on numerous examples in Figs.~\ref{fig:pull_figure} and \ref{fig:youtube_figure} and in our supplemental video.
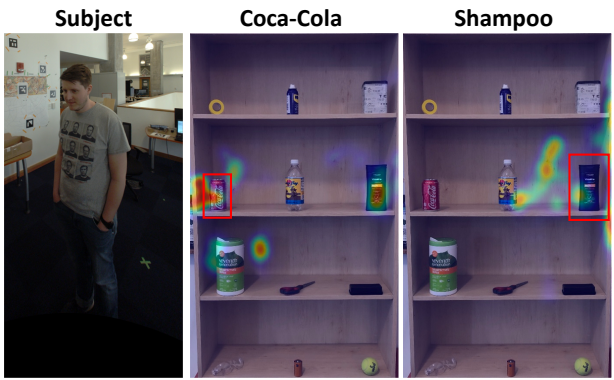
\myfigure{supermarket_figure}{\textbf{An example application}: we use Gaze360 to passively infer the attention of a customer as they browse products on a shelf, using video (left) from a camera next to the shelf (right).}

\textbf{Estimating attention in a supermarket}: To illustrate one possible application of Gaze360, we apply it to the task of predicting which objects are being looked at on a supermarket shelf, which is relevant for product-placement in stores. We recreate a supermarket shelf and ask subjects to look at various objects while self-reporting those objects. We record them with a camera next to the shelf, as shown in \refFig{supermarket_figure}. Despite a less than optimal view of the subject 
, we are able to predict which object is being looked at correctly $51\%$ of the time. Using a smartphone camera embedded directly in the shelf (so that the view of subjects is closer to frontal), the accuracy increases to $68\%$. The objects along the bottom shelves have highest error rate, as the eyes become almost fully occluded when looking downwards. Finally, we are able to produce a heatmap of customer attention, shown in \refFig{supermarket_figure}. While simple, this application demonstrates the flexibility of our system for use in a wide range of real-world applications. 

\section{Conclusion}

In this work, we introduced a novel approach to efficiently collect annotated gaze data at scale and used it to generate a large and diverse dataset, suitable for deep learning of 3D gaze from images and video. 
We presented a new temporal appearance-based gaze model using a novel loss function to estimate error quantiles.
Finally we demonstrated the value of (i) our dataset via careful cross-dataset performance comparison versus three existing 3D gaze datasets, and (ii) our model via application to unconstrained unseen imagery from YouTube videos.
It is our hope that by using our dataset and model, researchers across a range of fields will be able to better leverage gaze as a cue to improve vision-based understanding of human behavior.


\textbf{Acknowledgements.} 
Toyota Research Institute provided funds to assist the authors with their research but this article solely reflects the opinions and conclusions of its authors and not TRI or any other Toyota entity.

{\small
\bibliographystyle{ieee_fullname}
\bibliography{egbib}
}

\clearpage

 \end{document}

%% file: our-commands.tex
\newcommand{\degrees}{$^{\circ{}}$}

\DeclareGraphicsExtensions{.png,.pdf,.eps,.jpg}

\def\figurePath{figs/}
\def\myfigure#1#2{\begin{figure}[t!]\centering\includegraphics*[width = \linewidth]{\figurePath#1}\caption{#2}\label{fig:#1}\end{figure}}

\def\mycfigure#1#2{\begin{figure*}[t]\centering\includegraphics*[clip, width = \linewidth]{\figurePath#1}\caption{#2}\label{fig:#1}\end{figure*}}

\newcommand{\refSec}[1]{Sec.~\ref{sec:#1}}

\newcommand{\refFig}[1]{Fig.~\ref{fig:#1}}

\newcommand{\mycomment}[1]{}

\definecolor{darkred}{rgb}{0.6,0,0}
\definecolor{green}{rgb}{0.0,0.5,0}
\definecolor{blue}{rgb}{0,0,0.75}
\definecolor{orange}{rgb}{1,0.6,0.2}
\definecolor{red}{rgb}{1,0,0}

\soulregister\ref{7}
\soulregister\cite{7}
\soulregister\refFig{7}

\newcommand{\video}[1]{}

\newcommand{\mypm}{\mathbin{\smash{%
\raisebox{0.35ex}{%
            $\underset{\raisebox{0.5ex}{$\smash -$}}{\smash+}$%
            }%
        }%
    }%
}